\documentclass[conference]{IEEEtran}
\IEEEoverridecommandlockouts

\usepackage{cite}
\usepackage{amsmath,amssymb,amsfonts}
\usepackage{algorithmic}
\usepackage{graphicx}
\usepackage{textcomp}
\usepackage[dvipsnames]{xcolor}
\usepackage{comment}
\def\BibTeX{{\rm B\kern-.05em{\sc i\kern-.025em b}\kern-.08em
    T\kern-.1667em\lower.7ex\hbox{E}\kern-.125emX}}

\usepackage{booktabs}
\usepackage{url}

\usepackage{color}      
\usepackage{listings}   

\definecolor{codegreen}{rgb}{0,0.6,0}
\definecolor{codegray}{rgb}{0.5,0.5,0.5}
\definecolor{codepurple}{rgb}{0.58,0,0.82}
\definecolor{backcolour}{rgb}{0.95,0.95,0.92}
\definecolor{codecyan}{rgb}{0.0,0.2,1.0}

\lstdefinestyle{mystyle}{
    commentstyle=\textcolor{codegreen},
    keywordstyle=\color{codecyan},
    numberstyle=\tiny\color{codegray},
    stringstyle=\color{codepurple},
    basicstyle=\ttfamily\footnotesize,
    breakatwhitespace=false,    
    breaklines=true,    
    captionpos=b,    
    keepspaces=true,    
    numbers=left,    
    numbersep=2pt,  
    firstnumber=auto,
    numberblanklines=false,
    showspaces=false,
    showstringspaces=false,
    showtabs=false,
    tabsize=2
}
\usepackage{caption}
\usepackage{subcaption}

\usepackage{color}   

\usepackage{siunitx}

\newcommand{\quoteai}[1]{\emph{\textit{"#1"}}}
\newcommand{\prompt}[1]{\emph{\textcolor{ForestGreen}{#1}}}

\newcommand{\genai}{GenAI}
\newcommand{\chatgpt}{ChatGPT-4o}

\newcommand{\vscode}{VScode}

\begin{document}

\newcommand{\mytitle}{Case Study: Leveraging GenAI to Build AI-based
  Surrogates and Regressors for Modeling Radio Frequency Heating in Fusion Energy Science}

\title{\mytitle{} }


\author{E. Wes Bethel$^{1,4}$, Vianna Cramer$^1$, Alexander del Rio$^1$, Lothar Narins$^1$, Chris Pestano$^1$, Satvik Verma$^1$, \\
Erick Arias$^4$, Nicola Bertelli$^3$, Talita Perciano$^4$, Syun'ichi Shiraiwa$^3$, \'Alvaro S\'anchez Villar$^3$, \\
Greg Wallace$^2$, John C. Wright$^2$
\vspace{8pt}
  \\
$^1$ \textit{San Francisco State University, San Francisco CA, USA} \\
$^2$ \textit{Massachusetts Institute of Technology, Cambridge MA, USA} \\
$^3$ \textit{Princeton Plasma Physics Laboratory, Princeton NJ, USA} \\
$^4$ \textit{Lawrence Berkeley National Laboratory, Berkeley CA, USA} }

\maketitle

\begin{abstract}
This work presents a detailed case study on using Generative AI (GenAI) to develop AI surrogates for simulation models in fusion energy research. The scope includes the methodology, implementation, and results of using GenAI to assist in model development and optimization, comparing these results with previous manually developed models.
\end{abstract}

\begin{IEEEkeywords}
generative AI, surrogate modeling, regression, correlation analysis, model optimization, fusion energy science, AI-assisted code development and optimization
\end{IEEEkeywords}



\section{Introduction}

The need for real-time experiment predictions and control in fusion research has highlighted the limitations of traditional simulation codes like GENRAY, which do not run fast enough for real-time applications. There is significant interest in using AI as surrogates to approximate the results of full numerical physics computations within known or predictable error bounds, as demonstrated in our previous work~\cite{wallace2022towards}.

Currently, Generative AI (\genai{}) has gained substantial attention for its capabilities in creating images, videos, text, code, and even new AI models. Our focus in this work is to leverage \genai{} for creating numerical surrogates that approximate high-fidelity physics simulations.

This study explores using \genai{} as a \emph{well informed assistant} to aid in all steps of the model building, optimization, and evaluation process. 
In this context, we engage in a conversation with the \genai{} about each of these stages, and it provides suggestions about approach as well as initial code templates that we then adapt for use in our particular problem. 
This type of approach has proven effective for using AI in other contexts, such as education~\cite{bowen2024teaching}.

We leverage \genai{} to suggest approaches and provide code templates for the various stages of the model development pipeline, including exploratory data analysis; initial model development and evaluations; more extensive model optimization through k-fold cross-validation; final model training and evaluation. 
Being able to leverage \genai{} as part of this process has enabled us to reproduce most of our previous work in a shorter amount of development time and with marginally better model performance as well as perform a deeper exploration of model optimization strategies. 
Those include finding ways to reduce the number of input model features through different approaches like principal component analysis and by studying correlations between input features and output targets. 

The organization of this paper reflects the conversational nature of our interactions with \genai{}. 
We first reflect on background and previous work related to \genai{} and its role in developing software tools (\S\ref{sec:prevwork}).
Next, the narrative about approach and implementation begins with an overview then delves into each of the stages of model development, which is aimed at discovering the best parameters for each of the different types of models (\S\ref{sec:approach}).
Because preliminary results at earlier stages in the model development pipeline inform and impact decisions and action in later stages, these preliminary results are included as part of the narrative about implementation.
Once the best model parameters are identified, the final models undergo evaluation in terms of accuracy and computational speed (\S\ref{sec:results}).
\S\ref{sec:findings} provides observations and reflections about the results of the process of using \genai{} and about the nature of the quantitative and qualitative results of the study. 

The primary contributions of this paper are:

\begin{itemize}
\item A case study on using \genai{} to produce code templates for building AI surrogates for fusion physics simulations and for identifying and implementing strategies for model optimization.
\item  A comparison of AI-assisted codes in this work with human-generated codes in previous work in terms of the accuracy of the resulting models and computational requirements for model training and inference.
\item Use of \genai{} to identify potential pathways for improving model performance and code templates for implementing them along with some analysis results to support their use, or not, as the case may be. 
\item Anecdotal insights into out experience using \genai{} as a technical assistant. 
\end{itemize}

\section{Background and Previous Work}
\label{sec:prevwork}

\subsection{Use of AI models as Surrogates for Fusion Energy Simulations}

In fusion energy science, a tokamak is a machine that confines a burning plasma using a strong magnetic field~\cite{DOE:FES:Tokamak:2024}.
They are widely believed to be one of the most promising designs for a practical fusion reactors.
There is a robust international community of fusion energy scientists who study tokamak designs with an eye towards creating viable reactors for sustainably producing energy.

Among the challenges in the design and operation of a tokamak is the need to  understand what is happening inside the burning plasma along as well as the ability to manipulate it in various ways, such as reducing instabilities.
Radio Frequency (RF) systems such as lower hybrid current drive (LHCD) and high harmonic fast wave (HHFW) are well suited for use in tokamaks as they do not require line-of-sight access through radiation shields and they have a high degree of technology readiness. ~\cite{wallace2022towards}.
As such, being able to predict RF wave heating and current drive is essential for present-day real-time observation and control in fusion experiments and for modeling the designs of future devices.

As a practical matter, the computational methods used to predict RF wave heating and current drive have significant costs.
A single GENRAY/CQL3D simulation without radial diffusion of fast electrons requires 10s of minutes of wall-clock time to complete.
This runtime may be acceptable for some purposes, like offline modeling, but is much too slow for use in integrated modeling and real-time experimental control applications.

In previous work, Wallace et al., 2022~\cite{wallace2022towards} describe development of AI-based surrogate models to perform predictions of RF power absorption and current density profiles.
The results show a dramatic reduction in computational time, going from 10s of minutes to a few milliseconds once models are trained.
While the surrogate-based computations are not identical to those of the GENRAY/CQL3D code, that study quantifies differences using mean squared error. 

A significant amount of human effort was required by that time to generate a database of runs, followed by manual code generation to produce three different AI-based surrogates, and to measure their runtime and numerical accuracy performance.
In this case, significant effort is somewhat difficult to quantify, but can be characterized as the involvement of a team of four fusion physicists and four computer scientists working part-time over the period of about two years. 
In an effort to reduce time-to-solution for such an endeavor, the main objective for the work in this study is to explore the use of \genai{} for the purpose of producing similar AI surrogates, as well as to explore ways to improve model optimization and reduce computational cost of model training and inference.


\subsection{Using AI to Create AI Models and Software Tools}

Austin et al, 2021~\cite{Austin:2021} evaluate the efficacy of Large Language Models (LLMs) to synthesize code from natural language on two different datasets containing a total of about 24K programming problems. Their findings show solution rates between about 60-84\% accuracy for this particular problem set.

%
%
%

Denny et al. (2023)~\cite{Denny:Copilot:2023} investigate the limitations of GitHub Copilot in solving programming problems typically encountered in first-year computer science courses (CS1). Their study reveals that Copilot fails to generate a correct solution on the first attempt approximately 50\% of the time. They also highlight the importance of prompt engineering, noting that prompts enriched with detailed information, such as potential problem-solving strategies or pseudocode, tend to yield better results.

%
%
%

Idrisov et al. (2024)~\cite{Idrisov:2024} perform a comparative analysis of human- and AI-generated code across multiple metrics, including correctness, efficiency, and maintainability. Their study involves seven different generative AI-based systems, such as GitHub Copilot and CodeWhisperer, applied to solving LeetCode problems of varying difficulty. The results are mixed: some systems fail to produce correct solutions for any of the problems, while others generate correct code for certain problems but not others. Unlike their work, our study focuses on comparing the quality of numerical results produced by AI-generated code versus human-generated code for a specific scientific problem.

%

Ziegler et al. (2024)~\cite{Ziegler:2024} examine the impact of Copilot on programmer productivity. Their methodology includes both qualitative, self-reported data and quantitative metrics captured by Copilot, such as the acceptance rate of generated code. They find that the acceptance rate of Copilot's suggestions correlates more strongly with reported productivity than other persistence measures, such as the ratio of accepted completions that remain unchanged after a certain period. They propose that tools like Copilot facilitate progress toward users' goals by providing useful templates or starting points, which can be as beneficial as producing perfectly correct solutions. Our work incorporates both subjective measures (e.g., estimated duration to produce working, validated models) and objective measures (e.g., correctness of model predictions) to evaluate AI-based code generation in a scientific context.

These are but a few references of a growing body of work that is evaluating the ability of \genai{} systems to write code. 
For additional reading, please see ~\cite{Austin:2021} as well as Kotti et al., 2023~\cite{kotti:2023}, who summarize dozens of machine learning for software engineering studies from the years 2009--2022.

\section{Approach, Implementation, and Initial Results}
\label{sec:approach}

We begin with a brief overview of the process of creating AI models to provide a starting point for our implementation (\S\ref{sec:overview_model}). Even though we are focusing on regression models, many of these concepts are also applicable to classification models.
After a discussion of the source data, transformations, key assumptions and that influence aspects of our design and implementation (\S\ref{sec:source_data}), we dedicate a separate subsection to each of the three different AI-based regression methods; Random Forest Regression (\S\ref{sec:rfr}), Multilayer Perceptron Regression (\S\ref{sec:mlp}), and Gaussian Process Regression (\S\ref{sec:gpr}).
Included in the discussion of these methods is information about optimizing models through k-fold cross validation.
We then examine approaches for eliminating input features in an effort to reduce the computational cost for model training and inference (\S\ref{sec:reducing_features}).
We also provide a brief comparison of preliminary model results from these steps before moving on to final model training and evaluation in the next section.



\subsection{Overview of Model Development}
\label{sec:overview_model}

\begin{figure}[t!]
    \centering
    \includegraphics[width=0.45\textwidth]{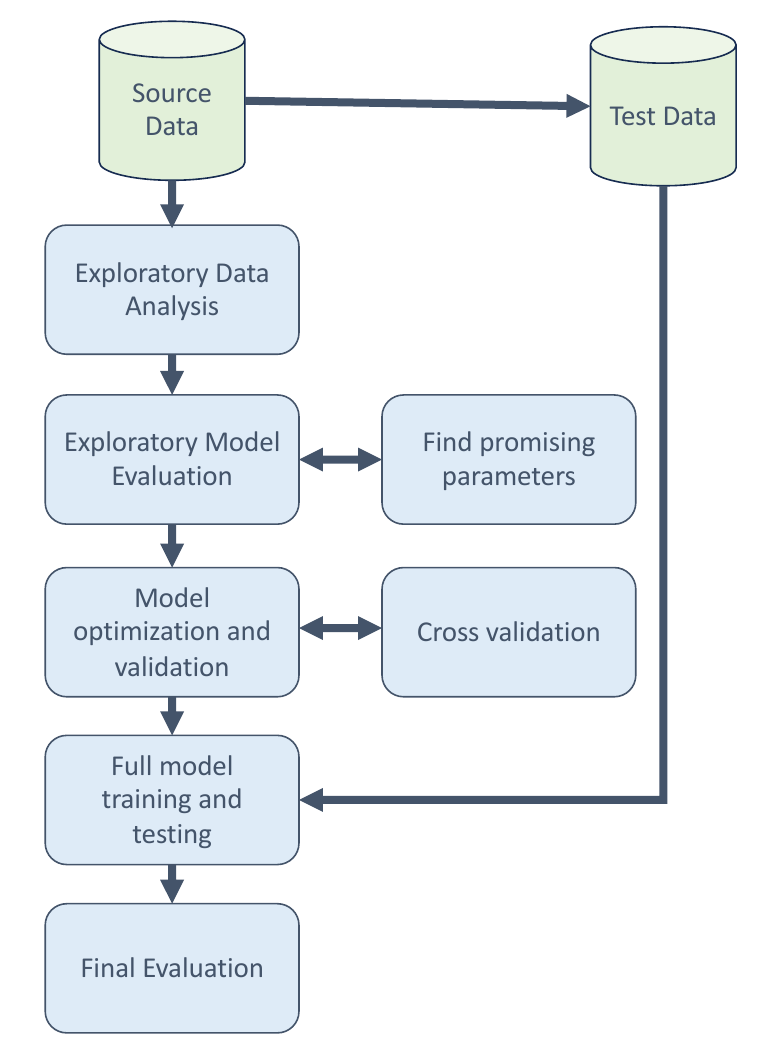}
    \caption{Overview and sequence of operations for model development, optimization, and evaluation.}
    \label{fig:flowchart}
\end{figure}

The process of developing an AI-based regressor involves several steps, including method selection, hyperparameter optimization, model training, and accuracy evaluation.
While the details of each step will vary depending upon the specific method being employed, the same overall set of themes is present for all methods.
Fig.~\ref{fig:flowchart} shows the sequence of processing steps we use in this study, and this sequence form the organization of material in the subsections that follow.

A first step likely includes exploratory data analysis for the purpose of deepening understanding of both the problem domain and characteristics of the data at hand~\cite{Muller:2016}. 
For our specific problem,  an initial exploration of the source data revealed that certain combinations of inputs resulted in code outputs that were not physically realistic. This can happen when the physical models do not perform well in certain ranges of input parameters. Ultimately, those data were filtered out so as to provide only physically realistic and meaningful inputs to model training~\cite{wallace2022towards}. 

A next step is to choose one or more AI models for the task at hand where consideration is given to 
strengths, weaknesses, and applicability to the problem at hand~\cite{Bishop:2006}.
For example, decision trees and random forests may be better suited for problems with non-linear relationships and interactions. 
Neural networks may perform well at capturing highly non-linear relationships between input features and output targets. 
Kernel-based methods might provide a better fit between a particular kernel combination choice and the underlying data. 
Typically, one might begin with initial model testing using default parameters to help identify promising methods.
In our case, we are focusing on use of three different methods that span three fundamentally different approaches: random forest, Gaussian process regression, and multilayer perceptron.
These three models were the focus of previous work where they were developed by hand prior to the prevalance of \genai{} tools~\cite{wallace2022towards}. 

An integral part of AI model development and testing is to evaluate its performance: how well do its predictions match the ground truth?
There are several different common metrics such as mean-squared error (MSE), $R^2$, mean absolute error (MAE) and others, each of which has particular strengths and weaknesses~\cite{hastie2009elements}. For our purposes, we are using MSE as a measure of model accuracy.

Given a particular model, a next step is adjusting  model parameters to improve model predictive performance. 
Each different model has its own unique set of parameters, and this process of parameter tuning is referred to as \emph{hyperparameter optimization}. 
Because many models' parameters cannot be directly estimated from the data~\cite{Kuhn_13}, the process of finding the optimal setting for model parameters may entail an iterative approach of model adjustment, evaluation, and testing.
Regular and systematic evaluation of the range of parameter values is known as a \emph{grid search} while a \emph{random search} uses a random values for parameters from a range and then makes adaptive search decisions based on model performance~\cite{Bergstra:2012}. 

A well established methodology for hyperparameter optimization is known is \emph{k-fold cross validation (CV)}~\cite{hastie2009elements}. 
The basic idea is to split a dataset into $K$
equally sized subsets, or folds. Then the model is trained $K$ times using $K-1$ folds for training and the remaining fold for validation. The process ensures that each fold is used for validation only once.
The average model performance, e.g., using MSE, is computed as the average of all individual model MSEs.

Complicating matters further is that each different model has a different set of parameters.
Random forest hyperparameters include the size of the forest, the maximum depth of the trees, and others~\cite{scikit-learn:rfr:web}.
Multilayer perceptron model hyperparameters include the number of network layers and connectivity, activation functions, learning rate, and regularization~\cite{haykin2009neural}.
GPR hyperparameters the choice of one or more kernel functions, kernel parameters like length scale and variance, and others~\cite{rasmussen2006gaussian}.



\subsection{Source Data Analysis and Preparation}
\label{sec:source_data}

\begin{figure}[h!]
    \centering
    \includegraphics[width=0.48\textwidth]{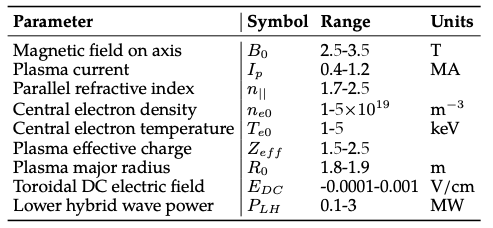}
    \caption{Parameters varied in the GENRAY/CQL3D database along with their ranges. Note the diverse dynamic range of these 9 parameters. Image source, Wallace et al., 2022~\cite{wallace2022towards}}
    \label{fig:genray-inputs}
\end{figure}

For this study, we are using the same dataset reported in Wallace et al., 2022 ~\cite{wallace2022towards}.
It consists of 13,347 records consisting of 9 simulation input variables and 2 output fields (current deposition and wave power profile) consisting of 23 variables each. 
The 9 input simulation variables, names, units, and numeric ranges are shown in Fig.~\ref{fig:genray-inputs}.
During original data preparation in 2022, the original 16K GENRAY outputs were filtered to eliminate data records from certain input parameter ranges known to correspond to output values that were physically not meaningful.

A common practice during supervised model training is to randomly divide the dataset into portions used for model training and model testing.
Because the 2022 study involved comparing 3 different models, to achieve consistency in which the same partitions of data are used for training and testing across the 3 different models, the study methodology included steps to perform data partitioning once then use those labeled partitions for all subsequent stages of the study.

In the 2022 study, a new column was added to the simulation data that identifies its "fold". The source data was first partitioned into an 80\%/20\% train/test split. The test split is held out from model training and optimization and used only in final model evaluation. 
The purpose of this "hold-out" fold of data is to validate the generalization ability of an AI model and ensuring that it will perform well on new, unseen data.
For the remaining 80\% of the simulation data, it was partitioned into 5 equal-size folds.
We adopt the same convention in our work here so as to achieve consistency in the data subsets used for k-fold CV as well as final model training and evaluation.

We use a common preprocessing step that normalizes the data values associated with the input features; 
their features and their ranges appear in Fig.~
\ref{fig:genray-inputs}.
We made use of the \texttt{sklearn.StandardScaler} method, which transforms data from its native range to a new range with a zero mean and unit variance. 
The reason for such resampling is two-fold. 
First, some methods like GPR kernels assume that features are centered around a mean of zero.
Another reason is the desire for the variances of all features to have the same magnitude so that features of larger numerical ranges do not dominate those with lower ranges during 
evaluation of objective functions~\cite{scikit-learn}.

\subsection{Random Forest Regression}
\label{sec:rfr}
 
The Random Forest model for classification or regression is an ensemble method that uses the averaging of predictions from many decision trees applied to random subsets of the training data~\cite{breiman2001random}.
Its advantages include robustness to overfitting through the averaging of predictions from multiple trees, as well as being effective with a larger number (dimension) of features. 

\begin{figure}[h!]
    \centering
\includegraphics[width=0.48\textwidth]{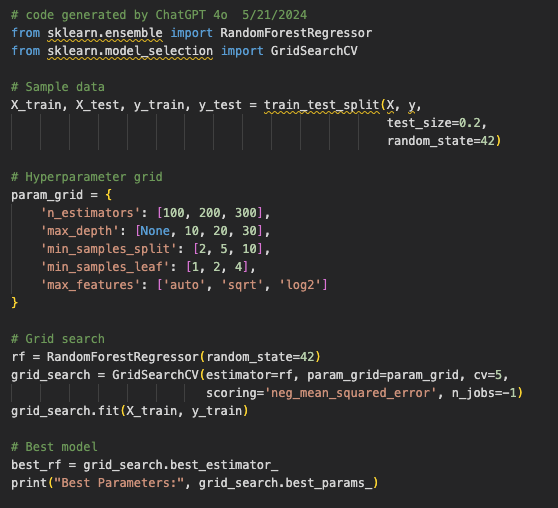}
    \caption{Source code generated by \chatgpt{} in response to a query asking for information about Random Forest Regressors.}
    \label{fig:rfr_source}
\end{figure}

Working with \chatgpt{}, we issued a prompt asking for \prompt{a description of a random forest regression along with key references}. In addition to some useful text-based information, it also provided the source code template shown in Fig.~\ref{fig:rfr_source}.
That source code includes some key processing steps, including: 
\begin{itemize}
    \item Dividing data into partitions to be used for model training and testing;
    \item Setting up a set of parameters and ranges to be used in  evaluating different model configurations;
    \item Code to perform a systematic (grid) search of the parameter space;
    \item Printing out the parameters of the best model
\end{itemize}

Starting with this code, we added our code to load the data file and separate features and targets, perform normalization as discussed in \S\ref{sec:source_data}, and extract one fold from the dataset for use in exploratory analysis.
The initial results were promising: the MSE was a bit higher than the mean MSE from 5-fold testing from 2022 as shown in Tab.~\ref{tab:results_mse} but still promising. 

It is important to note that in the code shown in Fig.~\ref{fig:rfr_source} that there is a grid of model parameters with initial values, and what follows is a "mini" k-fold CV combined with a random grid search courtesy of the \texttt{GridSearchCV} call.
Again, \chatgpt{} produced this code template in response to a query about RFR and their models.
While the grid search approach is common practice, 
changed from a grid search to a randomized search (\texttt{RandomizedSearchCV}), as it is known to run more quickly and to produce better results~\cite{Bergstra:2012}.

Next, we pursued further exploratory model evaluation  aimed at better understanding the RFR hyperparameters and their impacts.
This was an iterative process of asking questions of the \genai{}, tinkering with the code template by expanding the parameter grid variables and ranges, rerunning the code, and observing the changes in model MSE. 
We used the best model parameters from this portion of the study for subsequent hyperparameter optimization. 
The results of this early optimization step for all models is shown in Tab.~\ref{tab:exploratory_model_eval_mse}. 

Next, we engaged in a discussion with the \genai{} asking it to describe hyperparameter optimization in general terms as well as k-fold CV in particular. 
It shared the concept of \emph{nested cross-validation} that consists of an "inner loop", where a given test/train split is subject to the randomized parameter search using a CV strategy, and an "outer loop" where the different $K$ folds of data are rotated so that only one fold is used for validation.
This nested CV is an addition to the methodology from the 2022 study.
The resulting code has a structure shown in the Listing~\ref{listing:kfold_skeleton} is an abbreviated version of the actual code but that contains all key ideas.

\begin{lstlisting}[caption={Code skeleton for performing k-fold CV.},label={listing:kfold_skeleton}, name=stencil-core, float=h, style=mystyle,language=python]
# skeleton for k-fold processing
parm_grid = [ ... param names and ranges ... ]
best_model_parms = []  # for model parms in each fold
best_model_mse = [] # for model MSE in each fold
nfolds = 5
for fold in range(nfolds):
    X_train, y_train = extract_train(data, fold)
    X_test, y_test = extract_test(data, fold)
    
    # build the model
    model = RandomForestRegressor(random_state=42)
    
    # use a randomized search of model parm settings
    search = RandomizedSearchCV(...parameters...)
    
    # perform model fits to training data 
    search.fit(X_train, y_train)

    # obtain and retain params for best performing model
    best_model_parms.append(search.best_params_)
  

    # compute and retain MSE for the best model
    best_model = search.best_estimator_
    y_pred = best_model.predict(X_test)
    mse = mean_squared_error(y_test, y_pred)

    best_model_mse.append(mse)
\end{lstlisting}

From the k-fold CV process, we harvest the parameters of the best-performing model and use them to as parameters for the final, full-scale model training and testing.
For both stages, k-fold CV and final model, we collect runtime and MSE. 
These results are presented and discussed later in \S\ref{sec:results:mse_eval}.

\subsection{Multilayer Perceptron Regression}
\label{sec:mlp}

A Multilayer Perceptron (MLP) is a feed-forward artificial neural network composed of multiple layers of neurons (perceptrons) with activation functions.
It typically consists of three types of layers of nodes, an input layer, one or more hidden layers and an output layer. 
The MLP is a supervised learning algorithm that "trains" a neural network to map from inputs to outputs through a process of iteratively adjusting the layer weights and parameters backwards from the outputs to the inputs~\cite{rumelhart1986learning}.

\begin{figure}[h!]
    \centering
    \includegraphics[width=0.48\textwidth]{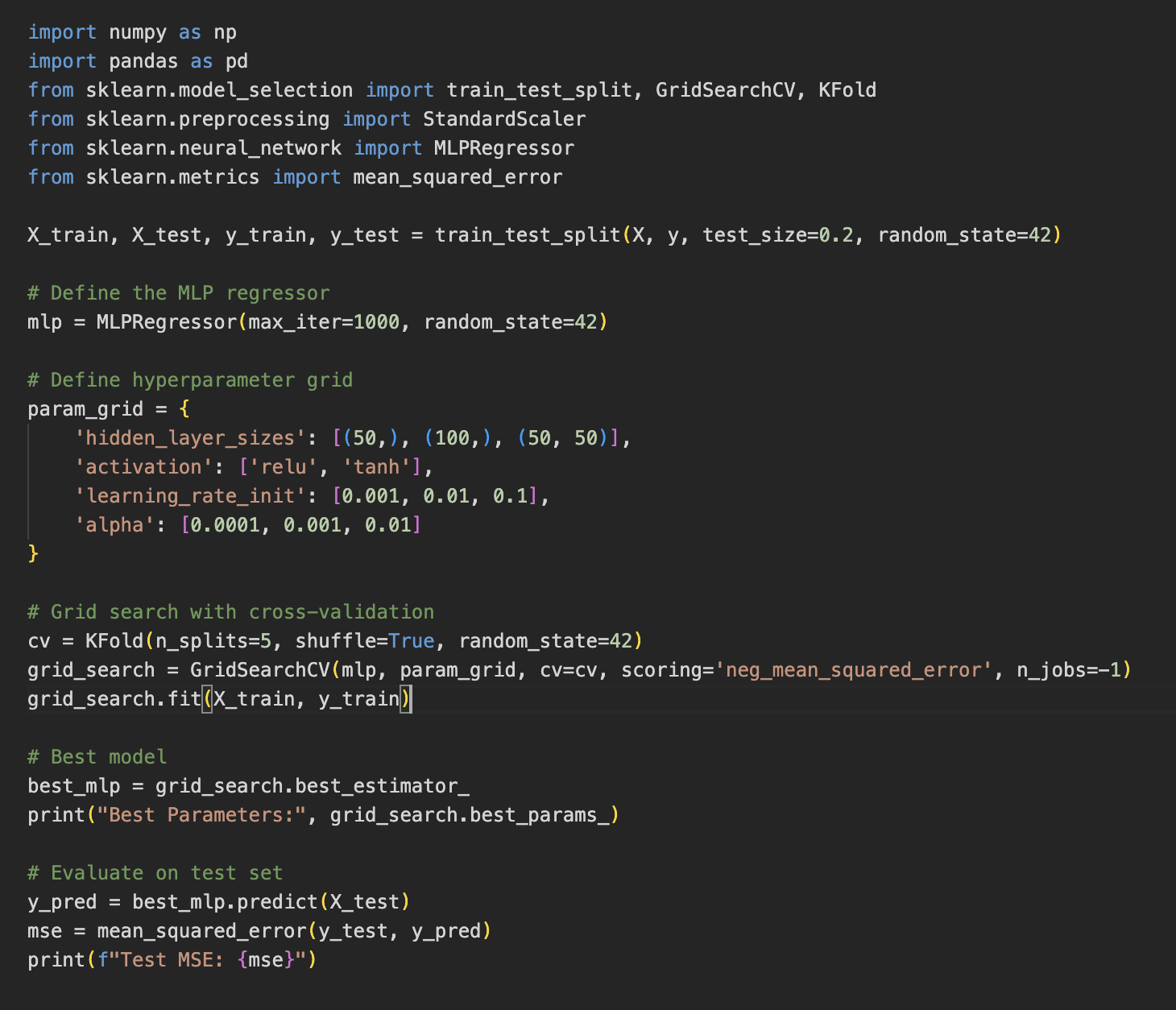}
    \caption{Source code generated by \chatgpt{} in response to a query asking for information about Multilayer Perceptron Regressors. }
    \label{fig:mlp_initial}
\end{figure}

Working with \chatgpt{}, we issued a prompt asking for \prompt{information about MLPs along with key references, and key steps in hyperparameter optimisation.}
In addition to useful text output,  \chatgpt{} generated an initial code template shown in Fig.~\ref{fig:mlp_initial}.

As in the case of the RFR code template, the MLP code template consists of the steps needed to build a basic MLP regressor starting with a simple set of model parameters and parameter ranges.
This starter code also performs a call to \texttt{GridSearchCV} to perform both a systematic search through the model parameter space that includes a 5-fold CV as part of the search.
The return value from the grid search is information about the model parameters that produced the best MSE results.

\begin{figure*}
     \centering
     \begin{subfigure}[b]{0.24\textwidth}
         \centering
         \includegraphics[width=\textwidth]{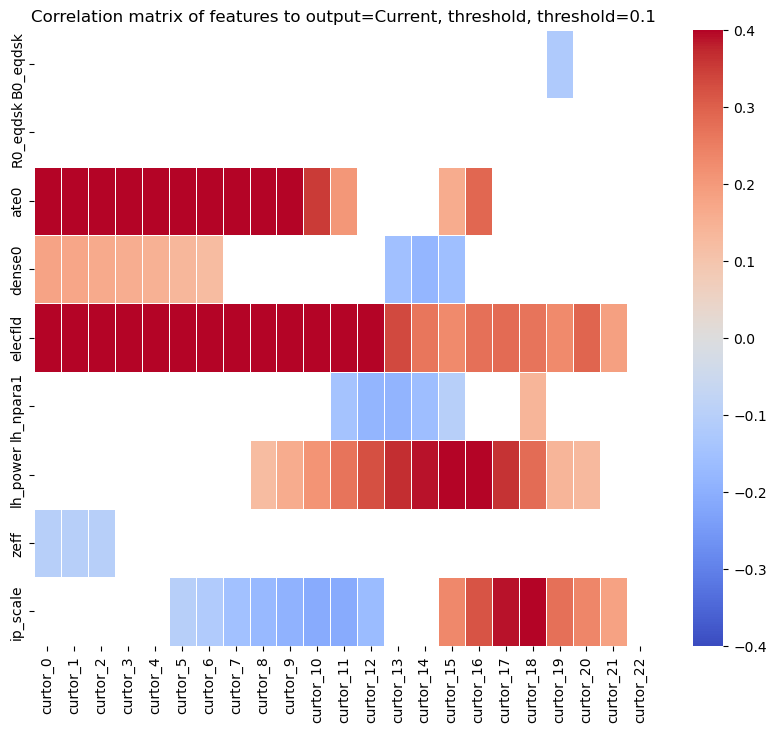}
         \label{fig:corr_map_curtor}
     \end{subfigure}
     \hfill
     \begin{subfigure}[b]{0.24\textwidth}
         \centering
         \includegraphics[width=\textwidth]{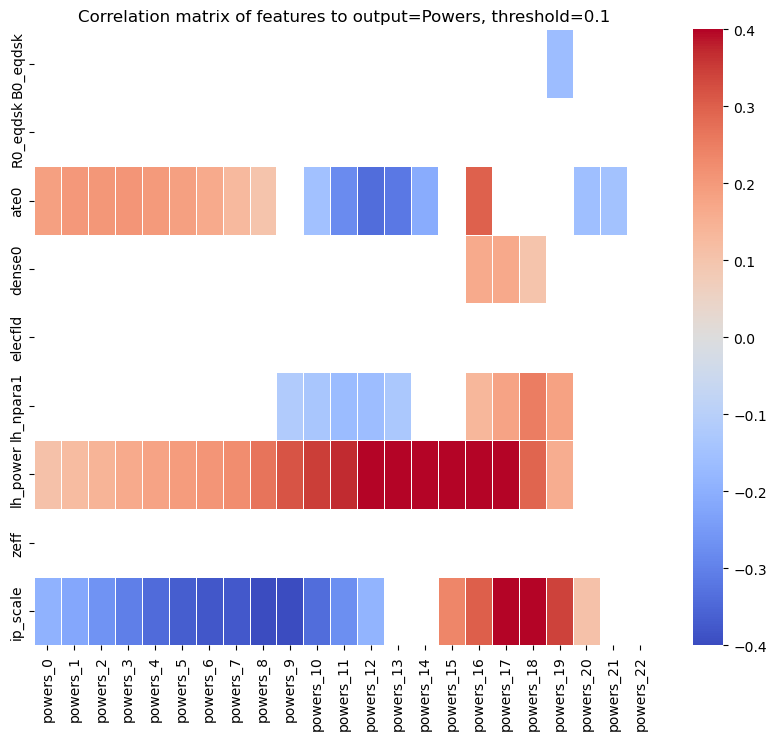}
         \label{fig:corr_map_powers}
     \end{subfigure}
   \hfill   
     \begin{subfigure}[b]{0.24\textwidth}
         \centering
         \includegraphics[width=\textwidth]{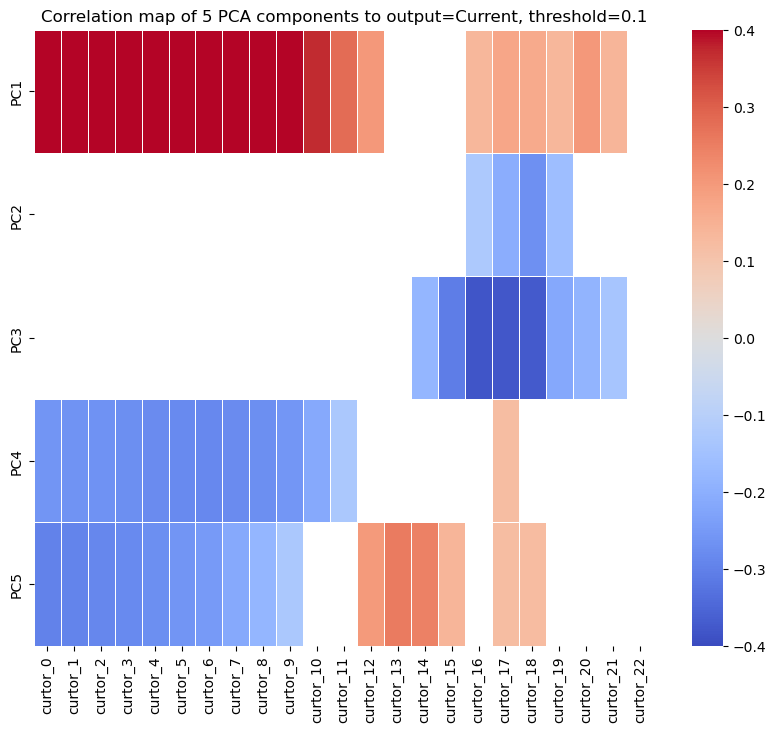}
         \label{fig:corr_map_pca5_curtor}
     \end{subfigure}
         \hfill
     \begin{subfigure}[b]{0.24\textwidth}
         \centering
         \includegraphics[width=\textwidth]{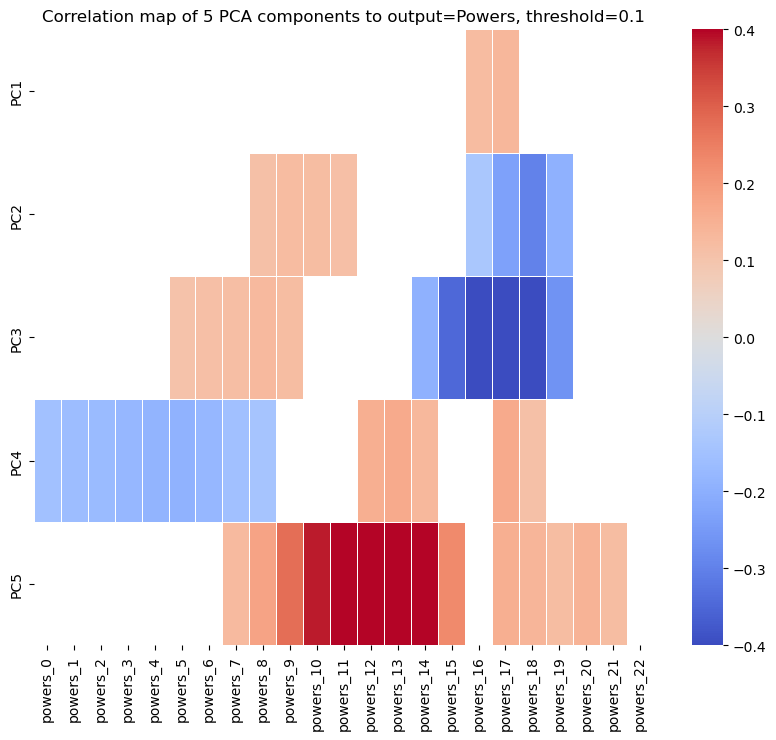}
         \label{fig:corr_map_pca5_powers}
     \end{subfigure}
    \caption{Correlation maps showing the relative strength of correlation between input features and output variables. Cells indicate the strength of the positive (red) or negative (blue) correlation between an input variable an an output feature. Cells where the absolute value of the correlation threshold $|t|<0.1$ are left blank. 
    The charts show correlations between input simulation variables for the variable Current (left), Powers (left-middle), and between 5 PCA components and the variable Current (right-middle) and Powers (right).
    }
        
\label{fig:corr_maps}

\end{figure*}

Like the RFR template example, we added our own code to load in one fold from the original dataset for the purpose of exploratory model analysis. 
We also changed from a grid search to a randomized search, as it is known to run more quickly and to produce better results~\cite{Bergstra:2012}.

Continued conversation with \chatgpt{} on the topic of model parameters led us to expand the range of parameter values.
For example, the code in Fig.~\ref{fig:mlp_initial} shows two single hidden layer configurations and one two-layer configuration. We explored several different permutations of network topologies (not shown here). 
Results from model optimization at this stage are reported in Table.~\ref{tab:exploratory_model_eval_mse}.

\subsection{Gaussian Processes Regression}
\label{sec:gpr}

In this study, we began to explore the same process for building GPR models as with the RFR and MLP models.
These codes have significant computational cost, as evidence by results from 2022~\cite{wallace2022towards} showing that the GPR methods have $10x$ the computational cost of RFR and MLP, but the resulting MSE is somewhere inbetween the two.
For the purpose of this study, we abandoned this particular line of investigation on the basis of computational costs.
Future work may examine finding ways to significantly reduce the computational costs.

GPR involves the inversion of an $N \times N$ covariance matrix (also known as the kernel matrix), where $N$ is the number of training samples~\cite{10.5555/1162254}.
The time complexity of inverting this matrix is $\mathcal{O}(N^3)$
and the memory complexity is $\mathcal{O}(N^2)$.
As the number of data points increases, the computational cost grows cubically and the memory cost grows quadratically, making GPR impractical for large datasets.

\subsection{Reducing the Number of Features}
\label{sec:reducing_features}

In an effort to reduce computational cost, we again consulted with \chatgpt{}.
Among its suggestions was one about reducing the number of input features, which in turn could reduce the computational cost of model training an inference.

Two different strategies came up in the conversation. 
One is to study the correlations between input features and output targets and perhaps exclude those features that do not have any significant impact on output features.
Another strategy is to perform Principal Component Analysis (PCA)~\cite{jolliffe2002principal} on the input features, and then perform model fitting on the resulting PCA components rather than the original input features. 

Since the key idea is the correlation between inputs and outputs, we iterated with \chatgpt{} to generate an initial code template for creating correlation maps.
We used that code first to create maps showing the correlation strength between input simulation variables (features) and outputs.
We iterated with \chatgpt{} to generate a code template that computes PCA components, and also created maps showing the correlation strength between 5 PCA components and the output variable.
All correlation maps are shown in Fig.~\ref{fig:corr_maps}. 

Based upon the results of this study, the suggested approach was to eliminate those features where no strong positive or negative correlation exists between the input feature and any output variable, where strongly correlated would be defined as $|t|>=0.1$.
From Fig.~\ref{fig:corr_maps}, for the variable Current, we can eliminate one input feature \texttt{R0\_eqdsk}, and for the variable Powers, we can eliminate three input features: \texttt{R0\_eqdsk}, \texttt{elecfld}, and \texttt{zeff}.
We modified versions of our RFR and MLP codes to use these reduced features.
There was some minor impact on the model in terms of computational rate and MSE both at the exploratory analysis phase (see Table~\ref{tab:exploratory_model_eval_mse} and during final model construction and analysis (\S\ref{sec:results:mse_eval}.

Looking at the correlation maps for the 5 PCA components in Fig.~\ref{fig:corr_maps}, none of the PCA components would appear to qualify for exclusion using the thresholding criteria of the absolute threshold value being smaller than some minimum, e.g., $|t|<0.1$.
Nonetheless PCA-1 and PCA-5 appear to have strong positive correlations for the variables Current and Powers, respectively.
Therefore, we created derivative codes for RFR and MLP that would perform PCA using 5 components, then used them as input features for model training.
The results shown in Table~\ref{tab:exploratory_model_eval_mse} reveal the model accuracy in terms of MSE is far inferior to that of the original simulation variables.
The fact that PCA did not produce advantageous results here is not a surprise: the original simulation features are more or less evenly distributed through their own parametric ranges owing to a Latin Hypercube sampling used for setting up the simulation runs~\cite{wallace2022towards}.
We did not further pursue the idea of using PCA components in this study.

\begin{table}[h!]
    \centering
     \def\arraystretch{1.25}
    \begin{tabular}{l c c}
   Method/features  & Current & Powers  \\
   \hline 
RFR, all features & 0.681 & 0.016 \\ 
RFR, reduced features & 0.688 & 0.015 \\ 
RFR, PCA-5 & 2.844 & 0.261 \\
\noalign{\vskip 4pt}    
MLP, all features & 0.339 & 0.014 \\ 
MLP, reduced features & 0.355 & 0.013 \\ 
MLP, PCA-5 & 2.779 & 0.025 \\
    \end{tabular}
    \caption{ Results of best model accuracy in terms of mean-squared error (MSE) from the exploratory model evaluation stage. The model parameters producing these MSE values were carried forward into the later model optimization and validation stage. The models built using PCA components as features had relatively poor performance, so this approach was not carried forward into later model development stages.}
    \label{tab:exploratory_model_eval_mse}
\end{table}
\section{Final Model Results}
\label{sec:results}

This section begins with a description of the hardware and software environment used for the model development and testing.
Next, we present quantitative findings from final model creation and evaluation, along with a comparison with earlier work.
We also examine the qualitative issue of whether or not development time was reduced compared to earlier efforts.

\subsection{Computational Environment -- Software}

In this study we leveraged two different \genai{} systems, Github's Copilot~\cite{github-copilot:2024} and OpenAI's \chatgpt{}~\cite{openai-chatgpt:2024}. 

When attempting to ascertain a version number for Copilot, we asked it directly and it said \quoteai{As an AI developed by OpenAI and GitHub, I don't have a specific version number like a traditional software application might. However, I'm based on the 
GPT-4 version of OpenAI's GPT models.}

Similarly, with \chatgpt{}, it responded \quoteai{I am an AI language model developed by OpenAI, based on the GPT-4 architecture. I do not have a specific version number like traditional software, but I am part of the GPT-4 series, which was released by OpenAI in March 2023. You can refer to me as "GPT-4".}

There are some troubling implications with respect to reproducibility given that there is no specific software version number to associate with a particular body of work, especially when that software is central as an assistant for concepts and code templates.

Other software that was leveraged as part of this work on both laptop/desktops and large-scale computational facilities includes:

\begin{itemize}
    \item sklearn~\cite{scikit-learn}, v1.5.0 for x86 Linux and MacOS
    \item GPflow~\cite{GPflow2017},  v2.9.1 for x86 Linux (Perlmutter@NERSC)
    \item Visual Studio Code (code)~\cite{vscode}, v1.89.1 on MacOS
    \item Python 3.10.0
\end{itemize}

\subsection{Computational Environment -- Hardware}

The team's general workflow is to use personal computers like laptops for interactions with Copilot through VScode. There was a diversity of platforms including x86 and Apple M1/M2 chipsets.
Interactions with the \genai{} systems, initial code development and testing occured on these personal platforms. 

For longer computational runs, the team made use of the Perlmutter system at the National Energy Research Scientific Computing Center (NERSC).
Perlmutter is a HPE (Hewlett Packard Enterprise) Cray EX supercomputer.
It is a heterogeneous system comprised of 3,072 CPU-only and 1,792 GPU-accelerated nodes. 

\begin{table*}[h!]
    \centering
    \begin{tabular}{l|cccc|cc}
    \toprule
    & \multicolumn{4}{c|}{5-Fold CV} & \multicolumn{2}{c}{Test (hold-out data)} \\
    & \multicolumn{2}{c}{Current} & \multicolumn{2}{c|}{Powers} & Current & Powers \\
    Method & $\mu(\text{MSE})$ & $\sigma(\text{MSE})$ & $\mu(\text{MSE})$ & $\sigma(\text{MSE})$ & $\mu(\text{MSE})$ & $\mu(\text{MSE})$\\
    \midrule
    MLP sklearn, all features   & $0.213$   & $0.018$   & $0.009$   & $0.0007$   & $0.210$ & $0.008$\\
    MLP sklearn, reduced features            & $0.258$   & $0.0129$   & $0.011$   & $0.0008$   & $0.251$ & $0.011$\\
    MLP TensorFlow (2022)           & $0.223$   & $0.015$   & $0.009$   & $0.0007$   & $0.227$ & $0.008$\\
\noalign{\vskip 4pt}  
    RFR sklearn, all features    & $0.562$   & $0.050$   & $0.016$   & $0.0007$   & $ 0.525$ &$0.015$ \\
    RFR sklearn, reduced features                  & $0.543$   & $0.046$   & $0.016$   & $0.0007$   & $0.508$ & $0.015$ \\
    RFR sklearn (2022)                   & $0.559$   & $0.045 $   & $0.016$   & $0.0007$   & $ 0.528$ &$0.015 $\\
    \vspace{4pt}
    \end{tabular}
    \caption{Evaluation of the three ML models using MSE as the performance metric, and comparing results of current study results with those from Wallace et al., 2022~\cite{wallace2022towards}. 
   For the 5-Fold cross-validation process we present the mean ($\mu$) and standard deviation ($\sigma$) of the MSE across all folds. The second column presents the prediction results of each final model trained using the full training data.}
    \label{tab:results_mse}
\end{table*}

\begin{table*}[h!]
    \centering
    \begin{tabular}{l|cc|cccc}
    \toprule
    & \multicolumn{2}{c|}{5-Fold CV}     & \multicolumn{4}{c}{Final model} \\
    & \multicolumn{2}{c|}{Training time (min)} & \multicolumn{2}{c}{Training time (min)} & \multicolumn{2}{c}{Inference time (ms)} \\
    Method & Current & Powers & Current & Powers & Current & Powers\\
    \midrule
    GENRAY/CQL3D                    & \multicolumn{1}{c}{--} & \multicolumn{1}{c|}{--} & -- & -- & \multicolumn{2}{c}{$O(10)$ (min)} \\
\noalign{\vskip 4pt}  
    MLP, all features         & \multicolumn{1}{r}{$149.32$} & \multicolumn{1}{r|}{$51.95$}   & \multicolumn{1}{r}{$1.65$}   & \multicolumn{1}{r}{$1.88$}    & \multicolumn{1}{r}{$0.003$} & {$0.004$} \\
    MLP, reduced features        & \multicolumn{1}{r}{$153.35$} & \multicolumn{1}{r|}{$60.43$}   & \multicolumn{1}{r}{$2.21$}   & \multicolumn{1}{r}{$1.89$}    & \multicolumn{1}{r}{$0.010$} & {$0.004$}\\
    MLP, (2022) all features         & \multicolumn{1}{r}{$71.7$} & \multicolumn{1}{r|}{$119.7$}   & \multicolumn{1}{r}{$14.40$}   & \multicolumn{1}{r}{$23.1$}    & \multicolumn{1}{r}{$0.800$} & {$1.690$}\\
\noalign{\vskip 4pt}  
    RFR, all features                   & \multicolumn{1}{r}{$20.49$}   &  \multicolumn{1}{r|}{$19.54$}  & \multicolumn{1}{r}{$1.52$}   &  \multicolumn{1}{r}{$1.57$} & \multicolumn{1}{r}{$0.480$} & {$0.488$} \\
    RFR, reduced features                   & \multicolumn{1}{r}{$15.90$}   &  \multicolumn{1}{r|}{$15.33$}  & \multicolumn{1}{r}{$1.52$}   &  \multicolumn{1}{r}{$0.59$} & \multicolumn{1}{r}{$0.533$} & {$0.256$} \\
    RFR, (2022) all features                   & \multicolumn{1}{r}{$3.2$}   &  \multicolumn{1}{r|}{$1.6$}  & \multicolumn{1}{r}{$9.10$}   &  \multicolumn{1}{r}{$3.9$} & \multicolumn{1}{r}{$0.930$} & {$0.750$} \\
    \end{tabular}
    \caption{Comparison of computation cost of performing 5-fold CV and final model training and testing. Model training times are measured in minutes, while inference times are measured in milliseconds. For comparison, the cost of one GENRAY "inference" run is $O(10)$ minutes compared to millisecond-scale timings for inference using AI-based surrogates. }
    \label{tab:results_timing}
\end{table*}

\subsection{Model Results and Comparison to Previous Implementation}
\label{sec:results:mse_eval}


The model accuracy results in terms of MSE for all models are shown in Table~\ref{tab:results_mse}.
These results include MSE from the four new models from this study: MLP versions for all features and reduced features, and RFR versions all features and reduced features applied to both Current and Powers variables.
For the sake of comparison, we are including MSE data for the two original models (MLP TensorFlow 2022, RFR sklearn 2022) applied to both Current and Powers variables.

In terms of overall model accuracy, both the MLP and RFR full-feature versions show minor accuracy improvement compared to their counterparts from 2022 for the Current variable and identical MSE for the Powers variable.
While both 2022 and present RFR codes are implemented using \texttt{sklearn}, the 2022 MLP is implemented using TensorFlow and the present MLP is implemented using \texttt{sklearn}.
The fact both RFR implementations use \texttt{sklearn}, use the same data, and virtually identical methodology leads us to expect the MSE will be nearly identical.
For MLP, the implementation differences along with differences in hyperparameter settings for the final model, due to differences in 5-fold CV outcomes, gives rise to a somewhat larger MSE difference for the Current variable.

Of the models that make use of reduced features, the value of the MLP metric MSE outcome is worse than (larger than) than the MSE metric of both 2022 and present MLP full-feature models.
On the other hand, the RFR reduced feature version show some MSE improvement compared to both 2022 and present versions for the Current variable. Results are identical for the powers variable for all RFR versions. 

For each of the different models, we accumulated the accuracy using MSE during both 5-fold training and final model construction. 
The results for all models in both stages of training appear in Table~\ref{tab:results_mse}, where they are juxtaposed with the corresponding model results from Wallace et al., 2022~\cite{wallace2022towards}.

The runtime costs for building models and performing inference workloads appear in Table~\ref{tab:results_timing}.
The present day models required significantly more runtime for 5-fold CV than those from 2022, most likely due to a greater number of different parameter configurations, a broader numeric range of parameter values, and the use of nested cross-validation.
This extra effort appears to have paid off in terms of the present day models requiring substantially less final model training time. 
Inference times for present day models also appear to be substantially less than for the 2022 models.

Comparing the present day reduced and all feature models, we see that the reduced feature models require more training time compared to the all feature models. The reverse is true for the RFR models.
This result is likely due to the data dependent performance differences in MLP versus RFR model optimization. The MLP optimizer may be working harder to converge when there are fewer features.

\begin{figure*}

     \centering
     \begin{subfigure}[b]{0.32\textwidth}
         \centering
         \includegraphics[width=\textwidth]{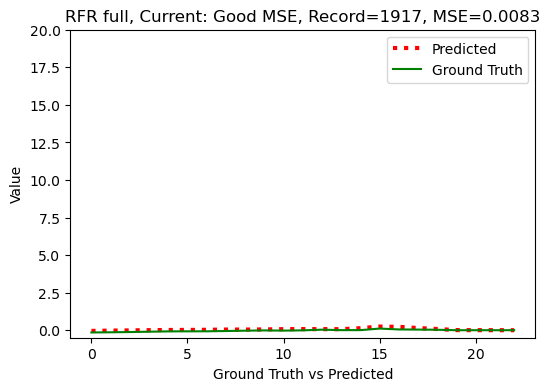}
         \label{fig:rfr_current_good}
     \end{subfigure}
     \hfill
       \begin{subfigure}[b]{0.32\textwidth}
         \centering
         \includegraphics[width=\textwidth]{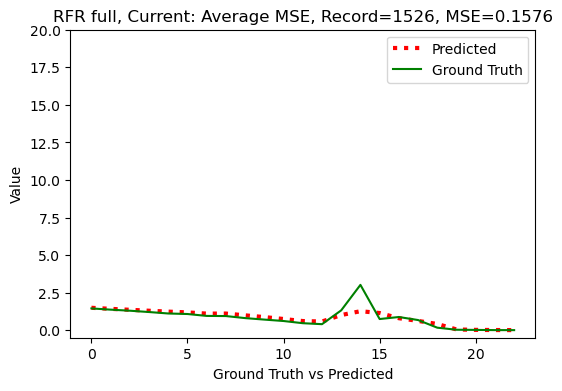}
         \label{fig:rfr_current_avg}
     \end{subfigure}
     \hfill
        \begin{subfigure}[b]{0.32\textwidth}
         \centering
         \includegraphics[width=\textwidth]{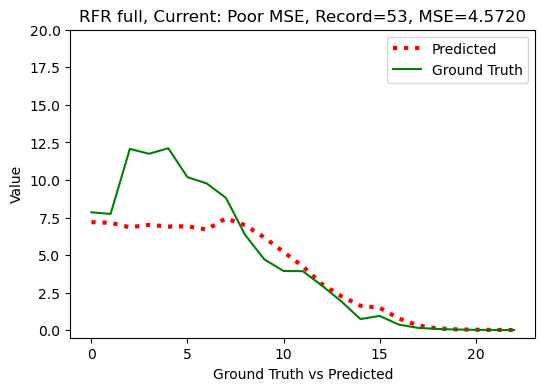}
         \label{fig:rfr_current_poor}
     \end{subfigure}
     \\
          \begin{subfigure}[b]{0.32\textwidth}
         \centering
         \includegraphics[width=\textwidth]{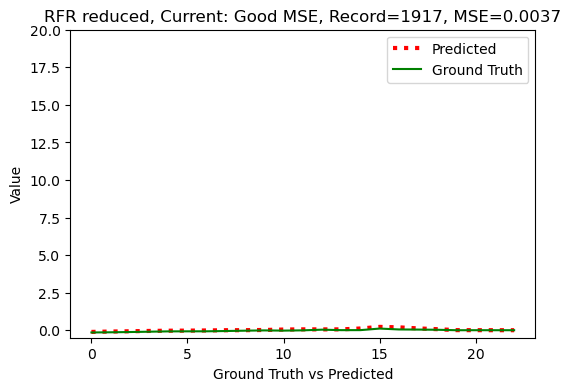}
         \label{fig:rfr_reduced_current_good}
     \end{subfigure}
     \hfill
       \begin{subfigure}[b]{0.32\textwidth}
         \centering
         \includegraphics[width=\textwidth]{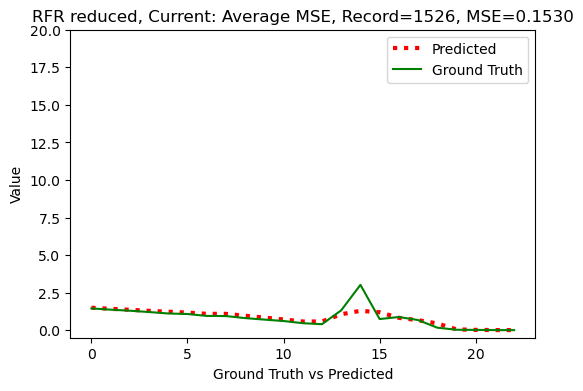}
         \label{fig:rfr_reduced_current_avg}
     \end{subfigure}
     \hfill
        \begin{subfigure}[b]{0.32\textwidth}
         \centering
         \includegraphics[width=\textwidth]{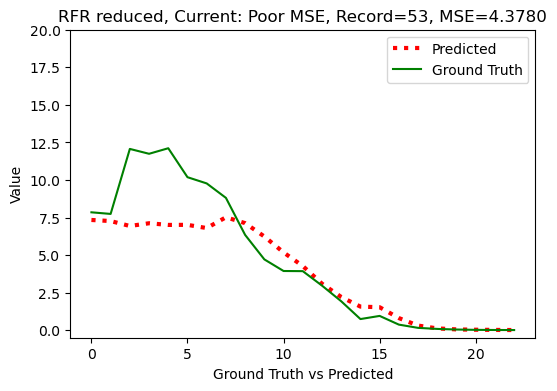}
         \label{fig:rfr_reduced_current_poor}
     \end{subfigure}
     \\
      \begin{subfigure}[b]{0.32\textwidth}
         \centering
         \includegraphics[width=\textwidth]{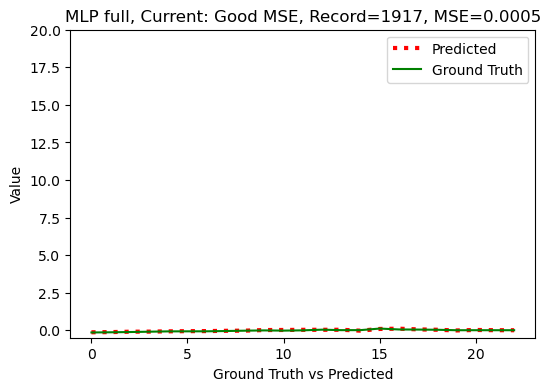}
         \label{fig:mlp_current_good}
     \end{subfigure}
     \hfill
     \begin{subfigure}[b]{0.32\textwidth}
         \centering
         \includegraphics[width=\textwidth]{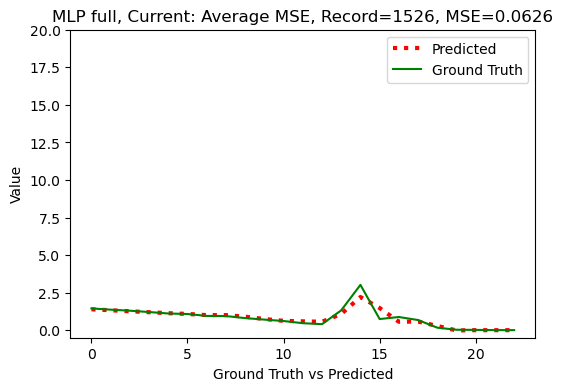}
         \label{fig:mlp_current_avg}
     \end{subfigure}
     \hfill
      \begin{subfigure}[b]{0.32\textwidth}
         \centering
         \includegraphics[width=\textwidth]{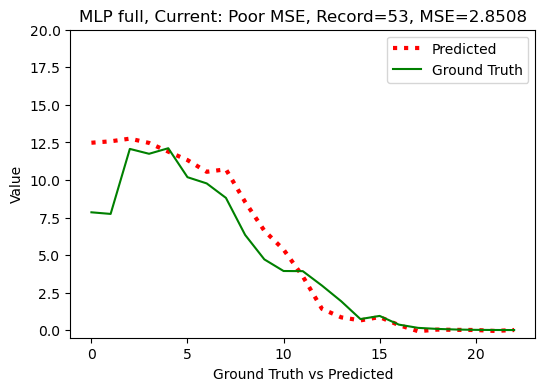}
         \label{fig:mlp_current_poor}
     \end{subfigure}
     \\
           \begin{subfigure}[b]{0.32\textwidth}
         \centering
         \includegraphics[width=\textwidth]{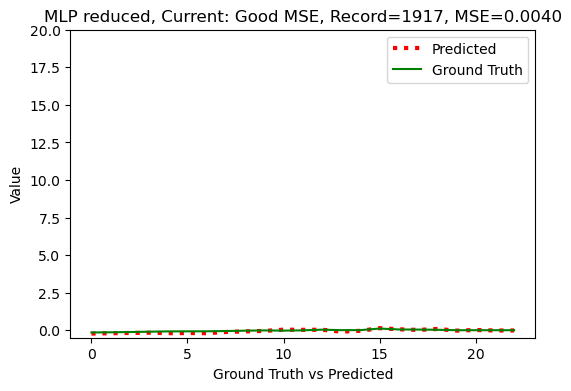}
         \label{fig:mlp_reduced_current_good}
     \end{subfigure}
     \hfill
     \begin{subfigure}[b]{0.32\textwidth}
         \centering
         \includegraphics[width=\textwidth]{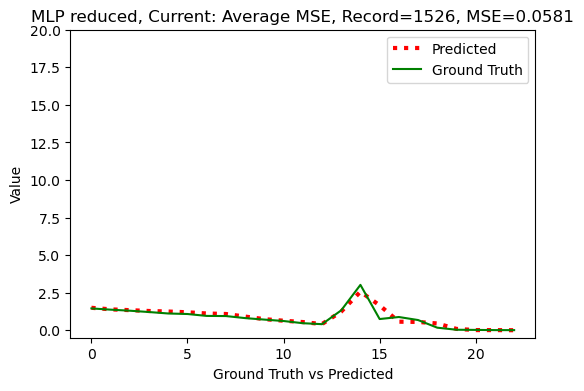}
         \label{fig:mlp_reduced_current_avg}
     \end{subfigure}
     \hfill
      \begin{subfigure}[b]{0.32\textwidth}
         \centering
         \includegraphics[width=\textwidth]{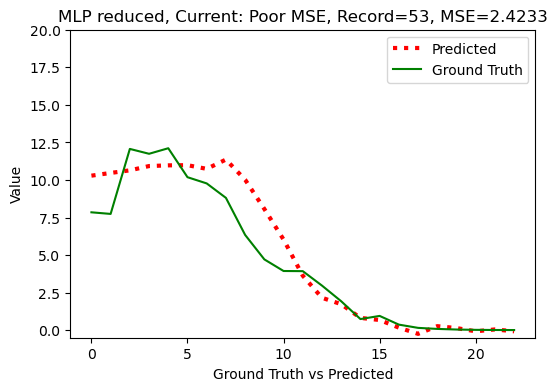}
         \label{fig:mlp_reduced_current_poor}
     \end{subfigure}
            \caption{For the variable current drive, comparison of model outputs for RFR with all features (first row), RFR with partial features (second row), MLP with all features (third row), and MLP with reduced features (bottom row). The left column shows typical "good" model accuracy, the middle column shows an "average" model accuracy, and the right column shows a "poor" model accuracy. Results for the variable \emph{powers} are similar and omitted for brevity. In these charts, the x-axis is the set of targets, or output values, produced by the model. The y-axis is the numerical value of each of these output values.
        }
        \label{fig:current_good_avg_poor}

\end{figure*}

In terms of visual comparison of model outputs, Fig.~\ref{fig:current_good_avg_poor} presents 12 plots comparing the results of ground truth against model predictions. There are four different models -- RFR all-features, RFR reduced-features, MLP all-features, MLP-reduced features -- and three different groups of "goodness of fit" based upon good, average, and poor MSE. The same simulation record is used for each of good, average, and poor fits to facilitate visual comparison amongst methods.
Note that this is one variable, Current. Due to space limitations, we are unable to show the other 12 plots for the Powers variable. They have similar visual characteristics.
These results are visually similar to those of the 2022 study~\cite{wallace2022towards}.



\subsection{Qualitative Results: Was Development Time Reduced?}

While some previous works have painstakingly measured developer performance to get a better feel for how \genai{} improves productivity (c.f.~\S\ref{sec:prevwork}, in this case we can offer some anecdotal observations. 

For the 2022 study~\cite{wallace2022towards}, the original team worked part-time over the course of two years on producing the first results. A significant amount of time was generating and validating the simulation database that is the input for model training. 
There were separate persons working on each of the three models, though each was concurrently working on multiple additional projects.
The 2022 study software also includes the creation of digital data artifacts that include the simulation database along with file-based versions of the trained models and Jupyter notebooks to load the train models and study data. 


The present effort began in February 2024 and these results represent about 4 months effort for a faculty and 3 students working part-time on the project.
All on the team are in agreement that we all feel more productive in terms of being able to interact with a capable \genai{} assistant who can present and summarize in-depth technical material along with code templates to get started.
Our own experience echoes that of other studies that conclude that \genai{} helps to improve coding and learning productivity.


\section{Findings and Discussion}
\label{sec:findings}

\genai{} may not do so well with data dependent operations. While Copilot and \chatgpt{} did a reasonably good job of producing general purpose code, it was not well suited for situations that require knowledge about the characteristics of a particular dataset or specific methods for working with those specific characteristics.

For example, when prompting Copilot to perform hyperparameter optimization for GPR, the resulting code exhibited lengthy runtimes and would not finish within the 6-hour limit for Juypter jobs. In response to a prompt about improving GPR runtime, \chatgpt{} suggested \quoteai{using a more efficient optimizer: The Scipy optimizer used in the code is a general-purpose optimizer. If you have some knowledge about the structure of the problem, you might be able to use a more efficient optimizer. However, this requires a deep understanding of the problem and the optimizers.}  

Results from \genai{} can be buggy. \chatgpt{} sometimes generates code with bugs, which we would repair using Copilot from inside \vscode{}. In addition, \chatgpt{} would sometimes provide citations to works that simply do not exist. All results from \genai{} should be subject to careful scrutiny and validation. 

It is well understood that there is bias in the output produced by \genai{} tools. This bias 
originates from several key factors related to the data used for training, the design of the models, and the algorithms that guide their learning. 
For example, \genai{} models are trained on vast datasets that are often scraped from the internet. These datasets can reflect the biases present in the real world, including cultural, social, and linguistic biases. 
In our studies, some of the initial code suggestions might not be the best for one reason or another. For example, both \chatgpt{} and Copilot always turn to a grid search for doing hyperparameter optimization. It is well known that a randomized search can often produce superior results in reduced runtime~\cite{Bergstra:2012}. We would often try out the initial grid search suggestion but then switch to a randomized search instead.


The GPR model was not included in this study because it was too computationally expensive. From the 2022 study, GPR was 10x more computationally expensive to train models compared to RFR and MLP, and its model accuracy was somewhere between RFR and MLP. While we did endeavor to find ways to reduce the computational runtime through a conversation with \chatgpt{}, none of the approaches provided satisfactory results. An open question is whether or not there are other approaches we have not found yet that would result in GPR being more competitive in terms of cost vs. accuracy for this particular problem. 


\section{Conclusion}



The overall objective for this work is leverage \genai{} for all stages of AI-based model conception, development, optimization, and evaluation. 
The results of this effort is compared both quantitatively and qualitatively with results from a 2022 study~\cite{wallace2022towards}.
The primary findings are that the current generation models have comparable performance in terms of accuracy, and there are some tradeoffs in terms of computational cost where more time spent optimizing current models also results in faster full model training and inference time. 

A key finding of this study is that use of \genai{} is far from being "turnkey". The \genai{} is best viewed as a capable assistant, and the information and code products it produces require careful scrutiny. In addition, the deeper the domain knowledge of the human conversing with the \genai{}, the better the quality of results. 


\section*{Acknowledgment}

In this work we made use of multiple \genai{} systems -- Github's Copilot~\cite{github-copilot:2024} and OpenAI's ChatGPT (4, and 4o)~\cite{openai-chatgpt:2024} -- for generating initial code samples and templates, for conversational exploration of technical topics that refined our approach, and for minor copyediting.

This work was supported by the Director, Office of Science, Office of Fusion Energy Sciences, of the U.S. Department of Energy under Contract Nos. DE-SC0021202 and DE-AC02-05CH11231, through the grant ‘Accelerating Radio Frequency Modeling Using Machine Learning’. 

This research used resources of the National Energy Research Scientific Computing Center (NERSC), a Department of Energy Office of Science User Facility using NERSC award FES-ERCAP-0028649.

The simulation runs producing results used for training data in this paper were performed on the MIT-PSFC partition of the Engaging cluster at the MGHPCC facility (\url{www.mghpcc.org}) which was funded by DoE grant number DE-FG02-91-ER54109.

\bibliographystyle{IEEEtran}
\bibliography{IEEEabrv,references}

\end{document}